\documentclass{article}

\PassOptionsToPackage{numbers, compress}{natbib}

\usepackage[final]{neurips_2022}
\usepackage{graphicx}
\usepackage{amsmath}
\usepackage{amssymb}
\usepackage{booktabs}
\usepackage{enumitem}

\usepackage[utf8]{inputenc} 
\usepackage[T1]{fontenc}    
\usepackage{hyperref}       
\usepackage{url}            
\usepackage{amsfonts}       
\usepackage{nicefrac}       
\usepackage{microtype}      
\usepackage{xcolor}         
\usepackage{comment}
\usepackage{caption}
\usepackage{tabularx}
\usepackage[export]{adjustbox}
\usepackage{amsmath}
\usepackage{array}

\usepackage[capitalize]{cleveref}
\crefname{section}{Sec.}{Secs.}
\crefname{table}{Tab.}{Tabs.}
\crefname{figure}{Fig.}{Figs.}
\crefname{equation}{Eq.}{Eqs.}

\newcolumntype{C}{ >{\centering\arraybackslash} m{1.5cm}}
\newcolumntype{D}{ >{\centering\arraybackslash} m{1.6cm}}
\newcolumntype{E}{ >{\centering\arraybackslash} m{2cm}}

\DeclareRobustCommand\onedot{\futurelet\@let@token\@onedot}
\def\onedot{. }

\title{\Large{Unpaired Image-to-Image Translation with Limited Data\\
to Reveal Subtle Phenotypes}}
\author{%
	Anis Bourou\\
	IBENS, ENS\\
	Université de Paris Cité\\
	\texttt{anis.bourou@ens.fr} \\
	\And
	Kévin Daupin \\
	Université de Paris Cité \\
	\texttt{kevindaupin@gmail.com} \\
	\AND
	Véronique Dubreuil \\
	Université de Paris Cité \\
	\texttt{veronique.dubreuil@u-paris.fr} \\
	\And
	Aurélie De Thonel\\
	Université de Paris Cité \\
	\texttt{aurelie.dethonel@u-paris.fr} \\
	\And
	Valérie Lallemand-Mezger \\
	Université de Paris \\
	\texttt{ valerie.mezger@u-paris.fr} \\
	\And
	Auguste Genovesio \\
	IBENS, ENS \\
	\texttt{auguste.genovesio@ens.psl.eu } \\
}

\begin{document}
\setcitestyle{square}

\maketitle
\begin{abstract}
Unpaired image translation aims at learning a mapping of images from a source domain to a target domain. Recently, these methods showed to be very useful in biological applications to display subtle phenotypic cell variations, otherwise invisible to the human eye. However, while most microscopy experiments remain limited in the number of images they can produce, current models require a large number of images to be trained. In this work, we present an improved CycleGAN architecture that employs self-supervised discriminators to alleviate the need for numerous images. We demonstrate, quantitatively and qualitatively that the proposed approach outperforms the CycleGAN baseline including when it is combined with differentiable augmentations. We also provide results obtained with small biological datasets on obvious and non-obvious cell phenotype variations demonstrating a straightforward application of this method.
\end{abstract}

\vspace{-5pt}
\section{Introduction}
\label{section intro}
\vspace{-5pt}
Image-to-image translation can be defined as transferring images from a source domain to a target domain while preserving the content representations. Image-to-image translation became popular in different fields such as autonomous driving cars\cite{car}, artistic style transfer\cite{arts} and more recently in biological imaging\cite{alixis}. Spotting differences in visual cell pheneotypes from medical and biological images has many applications in fundamental research, drug discovery and medicine. In microscopy experiments, hand-crafted image analysis can be used to measure the variation of phenotype produced by a perturbation when the last is visible. However, because cell-to- cell variability within an image often largely overlaps the cell-to-cell variability between phenotypes, the last is often invisible\cite{alixis}. This issue prevents researchers from observing and measuring subtle phenotypic differences between two images of cells.

The advent of GANs extended  the frontiers of image generation, GANs were also used in many approaches to build efficient image translation methods. In\cite{pix2pix}, the authors propose a supervised method based on conditional-GAN\cite{conditional} where the condition is the paired image target. Usually, obtaining paired images may be hard or even impossible. Therefore, unpaired image translation methods such as CycleGAN\cite{cycle}, DiscoGAN\cite{disco} and DualGAN\cite{dual} were proposed by exploiting the cycle-consistency constraint\cite{cycle}.

A common limitation with the methods cited above is that they all require a large number of images in both the source and the target domains to be trained effectively. Several strategies were proposed to tackle this issue \cite{tgan,cross-domain,funit}. In general, these methods assume the existence of a large image dataset from a close domain that can be exploited using different strategies. However, this assumption is not often met. In bio-imagery, for instance, it is hard to get images that resemble the images of organoids of interest while they cannot be produced by thousands.

Self-supervised learning (SSL) \cite{ssl1,ssl2} is a paradigm that tries to overcome the problem of the lack of labelled data by obtaining a supervision signal from the data itself in order to learn a richer representation. Recently, several works proposed to leverage SSL to solve tasks in setting where little training data is available\cite{boost,slefgan,fastgan}

In this work, we present \textbf{Self-Supervised CycleGAN}(SCGAN), a model based on the cycle-consistency constraint\cite{cycle} and self-supervised learning for an efficient image-to-image translation with limited data.

\setlength{\tabcolsep}{2pt}
\section{Proposed Method}

Given a source domain $X$ and a target domain $Y$, the goal of an image translation model is to learn a mapping $G: X \xrightarrow{} Y$ such that the output $\hat{y}=G(x)$, $x \in X$, is indistinguishable from images 
$y \in Y$. Our method (SCGAN) consists of two generators and two discriminators. The efficiency of the proposed approach relies mainly on the fact that we use self-supervised discriminators \cite{fastgan}. In the following sections, we describe the architectures and the losses we use to achieve satisfactory unpaired image translation with a low number of images.

\paragraph{Generator}
For the generators, we adopt the architecture used in \cite{cycle}. The model contains three convolutions followed by 9 residual blocks\cite{residual}, then we use two fractionally-strided convolutions and finally one convolution that maps the features to RGB. We also use instance normalization\cite{instance_norm}.
\paragraph{Self-supervised discriminator}
For the discriminator, we rely on the architecture proposed in \cite{fastgan}, where a strong regularization for the discriminator is provided through a self-supervised learning strategy. Unlike regular self-supervised approaches, where transformations of the same image are pushed to produce the same representation, this approach generates a self-supervised signal by learning to reconstruct a degraded version of the image. In this configuration the discriminator acts as an encoder trained with small decoders. These are simple networks made of four conv-layers which are jointly optimized with the discriminator on a simple reconstruction loss using the real samples only. For $D_X$, the adversarial discriminator for images from domain $X$, the ssl loss reads: 

\begin{equation}
     L_{ssl}(D_X) = \underset{{x\sim\mathbb{p}_{data(x)}}}{\mathbb{E}}\|Dec(T'(Enc(x)))-T(x)\| 
    \label{eqn:ssl}
\end{equation}
where $Enc$ produces an intermediate feature map in $D$ , the function $T'$ is a simple degradation process (such as crop) on $x$ and $T$ is a similar degradation process on the internal representation provided by $Enc$. Finally, $Dec$ is a decoder from the degraded internal representation to the
degraded image $T(x)$. In this setting, the discriminator is encouraged to extract a more comprehensive and useful representation yet using a lower number of data.

\paragraph{Training losses}
At training time, we use the least square loss introduced in LSGAN \cite{least_square} as it provides a more stable training by mitigating the vanishing gradient and mode collapse issues. The two generators and the two discriminators minimize the following losses, here for $G$:
\begin{equation}
    L_G = \underset{x\sim\mathbb{P}_X}{\mathbb{E}}[(D_Y(G(x))-1)^2]
\end{equation}
\begin{equation}
    L_{D_Y} = \underset{y\sim\mathbb{P}_Y}{\mathbb{E}}[(D_Y(y)-1)^2] + 
         \underset{x\sim\mathbb{P}_X}{\mathbb{E}}[(D_Y(G(x)))^2]+ L_{ssl}(D_Y)
\end{equation}
where $D_Y$ denotes the adversarial discriminator for images from domain $Y$ and $L_{ssl}(D_Y)$ is the self supervised loss described in the previous section. $L_F$ for the generator F and $L_{D_X}$ for the adversarial discriminator $D_X$ are trained similarly and altogether the adversarial loss is:
\begin{equation}
    L_{adv} = L_G + L_{D_Y} + L_F + L_{D_X}
\end{equation}
In addition, the cycle-consistency constraint ensures that an image $x$ from domain $X$ consecutively transformed by the two generators reconstructs the original image $x$, i.e., $F(G(x)) \approx x$. Similarly, for each image $y$ from domain $Y$, $G(F(y))\approx y$. This constraint is enforced by the following loss:

\begin{equation}
    L_{cyc} = \underset{x\sim\mathbb{p}_X}{\mathbb{E}}[\|F(G(x))-x\|_1] + \underset{y\sim\mathbb{p}_Y}{\mathbb{E}}[\|G(F(y))-y\|_1]
\end{equation}

In addition to the adversarial and the cycle consistency losses, it was shown in \cite{cycle} that adding an identity loss \cite{identity} can enhance the performance of the CycleGAN. This loss ensures that transforming an image from a domain to the same domain will not modify it. It is defined as follows: 

\begin{equation}
    L_{id} = \underset{x\sim\mathbb{p}_X}{\mathbb{E}}[\|F(x)-x\|_1] + \underset{y\sim\mathbb{p}_Y}{\mathbb{E}}[\|G(y)-y\|_1]
\end{equation}

The full objective loss to minimize is the summation of the three losses: the adversarial, the cyclic and the identity, it is given as follows:

\begin{equation}
    L_{total} = L_{adv} + \lambda_1 L_{cyc} + \lambda_2 L_{id} 
\end{equation}

\vspace{-15pt}

\section{Experiment}
\label{sec:typestyle}

\subsection{Datasets and training}

\paragraph{Horse2Zebra.} The images of horses and zebras were downloaded from ImageNet\cite{imagenet}. The images were scaled to 256 × 256 pixels. The training set size of each class is: 939 (horse) and 1177 (zebra).

\paragraph{BBBC021.} Microscopy images of MCF-7 cancer cells untreated (DMSO) and treated for 24h with Cytochalasin B at high dosages (30$\mu$M). Training was performed on 200 images from each condition.

\paragraph{Organoids.} Microscopy images of neural organoids (mini-brains) induced from the stem cells of a rare neuro-developmental disorder. We used three marker: ZO1 for the cell-to-cell junctions, phosphohistone H3 for the dividing cells and DAPI for the nuclei. Training was performed on 56 images of healthy organoids and 83 images of diseased organoids. 

\begin{figure}[htb]

\begin{minipage}[b]{.48\linewidth}
  \centering
  \centerline{\includegraphics[width=4.0cm]{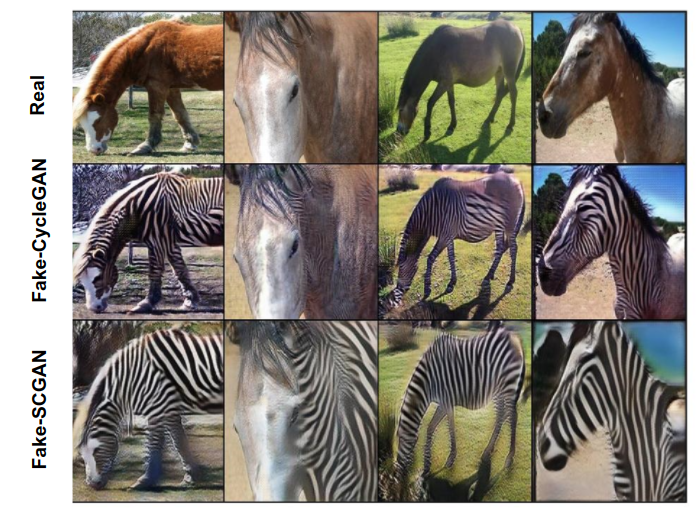}}
   \centerline{}\medskip
   \caption{Translating horse images to zebra images using only 40 percent of the dataset. We can see that our method outperforms CycleGAN.}
   \label{fig:horse}
\end{minipage}
\hfill
\begin{minipage}[b]{0.48\linewidth}
  \centering
  \centerline{\includegraphics[width=4.0cm]{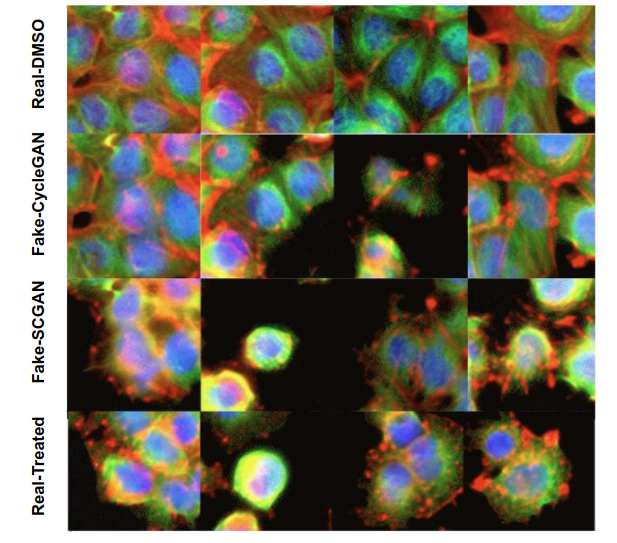}}
  \caption{Translated images of untreated (DMSO) cell (first row) into drug treated cells using CycleGAN and SCGAN (2nd and 3rd rows respectively, we can see that the translation made by SCGAN is close to the real images of treated cells (4th row).}
  \label{fig:frame}
\end{minipage}

\end{figure}

\paragraph{Training.} For all the experiments, we set $\lambda_1 = 10$ and $\lambda_2 = 0.5$. We used the Adam optimizer with a batch size of 8 and a learning rate of $0.0002$. This rate was decayed linearly once every $100$ epochs. We used FID\cite{fid} and KID\cite{KID} to compare the image distributions, the lower the better. We compared SCGAN to CycleGAN with and without adding a differentiable augmentation module. We trained the models using different percentages of the horse2zebra dataset. We also trained it on biological images to illustrate the portability of our approach.

\subsection{Results}
\begin{figure}[htb]

\begin{minipage}[b]{.48\linewidth}
  \centering
  \centerline{\includegraphics[width=5.0cm]{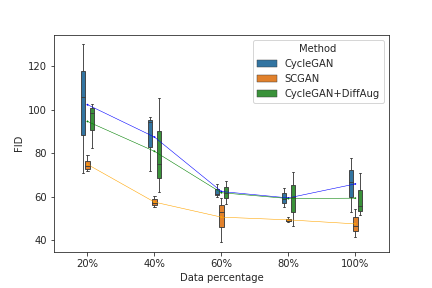}}
  \centerline{Comparing KID}\medskip
\end{minipage}
\hfill
\begin{minipage}[b]{0.48\linewidth}
  \centering
  \centerline{\includegraphics[width=5.0cm]{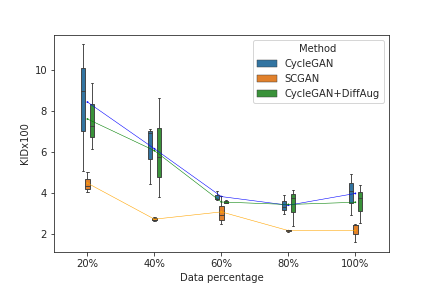}}
  \centerline{Comparing FID}\medskip
\end{minipage}
\caption{Comparing the CycleGAN, CycleGAN+DiffAug and SCGAN in terms of (a) FID and (b) KID using different percentages of the horse2zebra dataset. With each data perecentage, we train three times our model, the boxplots are plotted using KID and FID scores computed for each run.}
\label{fig:resImg}
\end{figure}

The FID curves in fig. \ref{fig:resImg} show the effectiveness of our method compared to the standard CycleGAN. The results based on FID and KID scores show that SCGAN improves the translation with any quantity of data for the horse2zebra dataset. While differentiable augmentations\cite{diff_aug} can slightly improve over CycleGAN, our method remains significantly better. Qualitatively, our method achieves a better translation when trained with only 40 percent of the horse2zebra dataset (fig. \ref{fig:horse}).

In fig.\ref{fig:frame}, we show that we also achieved better translation on a biological dataset where untreated cells (DMSO) were treated with a high concentration of compound (Cytochalasin B) which produces an obvious change in phenotypes with as little as 200 images from each class. Quantitatively, we obtained \textbf{FID = 77.84 and KIDx100 = 6.53} with CycleGAN and  \textbf{FID = 55.36 and KIDx100 = 2.56} with our approach. This indicates that our method outperforms CycleGAN by a large margin.

\begin{figure}[t]
\begin{center}
   \includegraphics[width=0.6\linewidth]{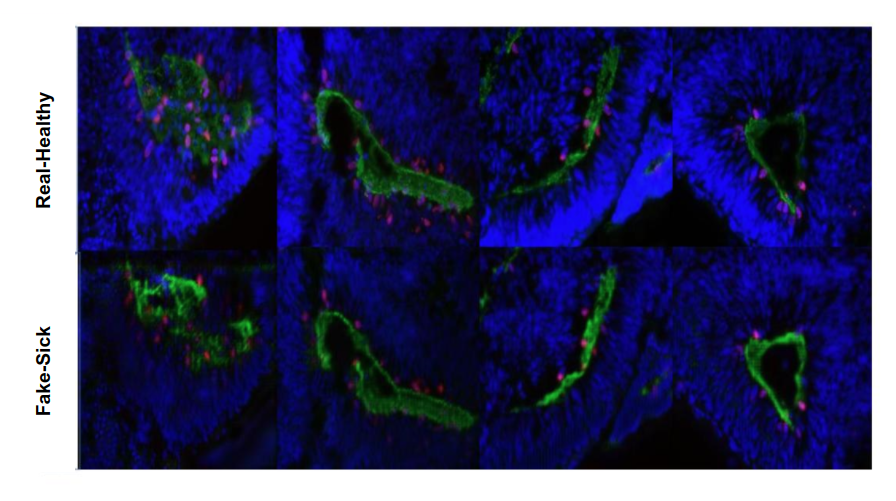}
 \caption{In the first row we have the real images of the healthy organoids and in the second row we have their translation to the other class.} 
 \label{img:drug}
\end{center}\vspace*{-\baselineskip}

\label{fig:org}
\end{figure}

In fig. \ref{img:drug}, we then applied our approach on invisible phenotypic changes between two conditions on organoids. The translated images showcase some changes: 1) the intensity of the blue marker has diminished indicating a decrease in the number of neural cells attained with the syndrome. There are also fewer red cells in the translated images compared to the real ones indicating a decrease in cell divisions in the the diseased cells. In order to validate these subtle differences, further biological experiments should of course be conducted.

\section{Conclusion}
In this work, we presented SCGAN, a self-supervised cycle-consistency based method for unpaired image translation with limited size datasets. We showed through experiments that our method outperforms the standard CycleGAN with and without augmentation strategies used to overcome data scarcity. Furthermore, we showcased the importance of self-supervised learning as an efficient strategy to mitigate the need for large size datasets. Finally, we showed how image translation can be applied in biology to identify subtle phenotypes when the number of the available images is limited. It can be used to guide the intuition of the experts to understand subtle biological processes or identify new therapeutic biomarkers.

\section{Acknowledgments}
\label{sec:acknowledgments}
This work was supported by ANR–10–LABX–54 MEMOLIFE and ANR–10 IDEX 0001
–02 PSL* Université Paris, ANR Visualpseudotime and was granted access
to the HPC resources of IDRIS under the allocation 2020-
AD011011495 made by GENCI.

{\small{
\bibliographystyle{ieee_fullname}
\bibliography{egbib}}
}

\end{document}